# 3D Object Detection Combining Semantic and Geometric Features from Point Clouds

**Hao Peng[1], Guofeng Tong[1,*] Zheng Li[1], Yaqi Wang[1] Yuyuan Shao[1]**

[1]College of Information Science and Engineering, Northeastern University, Shenyang 110819, Liaoning, China.
[*] Corresponding author.
Email:
apengh@126.com (Hao Peng)

**Abstract**
In this paper, we investigate the combination of voxel-based methods and point-based methods, and propose a novel end-to-end two-stage 3D object detector named SGNet for point clouds scenes. The voxel-based methods voxelize the scene to regular grids, which can be processed with the current advanced feature learning frameworks based on convolutional layers for semantic feature learning. Whereas the point-based methods can better extract the geometric feature of the point due to the coordinate reservations. The combination of the two is an effective solution for 3D object detection from point clouds. However, most current methods use a voxel-based detection head with anchors for final classification and localization. Although the preset anchors cover the entire scene, it is not suitable for point clouds detection tasks with larger scenes and multiple categories due to the limitation of voxel size. In this paper, we propose a voxel-to-point module (VTPM) that captures semantic and geometric features. The VTPM is a Voxel-Point-Based Module that finally implements 3D object detection in point space, which is more conducive to the detection of small-size objects and avoids the presets of anchors in inference stage. In addition, a Confidence Adjustment Module (CAM) with the center-boundary-aware confidence attention is proposed to solve the misalignment between the predicted confidence and proposals in the regions of the interest (RoI) selection. The SGNet proposed in this paper has achieved state-of-the-art results for 3D object detection in the KITTI dataset, especially in the detection of small-size objects such as cyclists. Actually, as of September 19, 2021, for KITTI dataset, SGNet ranked ***1st*** in 3D and BEV detection on cyclists with easy difficulty level, and ***2nd*** in the 3D detection of moderate cyclists.

**Keywords**
3D object detection, point clouds, voxel-to-point, semantic, geometric.

# 1 Introduction

3D object detection is an important task of scene understanding in computer vision and autonomous driving, which can perceive the category and location of the object interested in the scene [1]. Lidar can accurately reproduce 3D spatial information, making it widely used in the field of autonomous driving. And, 3D object detection based on point clouds has been laid more and more emphasis in recent years.

Point-based, voxel-based, and the combination of the two are the mainstream methods for 3D object detection from point clouds. The point-based methods [2, 3, 4] can retain the fine-grained spatial information of the scene, which is conducive to capturing the geometric characteristics of the object. And, many invalid detections in the space of the scene are avoided. However, the point-based method needs to perform a neighborhood search [5, 6] when learning the semantic features of a local area, which increases the inference time of the model. The voxel-based methods [7, 8, 9, 10, 11] divide the entire space into regular grids, and then project them to BEV and use convolutional neural networks to learn the semantic features of the scene. These methods have strict requirements on the size of the grid which is an important factor for balancing the amount of calculation and accuracy. For example, although the large-size grid makes the inference speed fast, it brings obstacles to the detection of small-size target objects. Moreover, the voxel-based method is often accompanied by the preset of various types of anchors. Due to complexity considerations, this method is not suitable for scenes with many categories and large sizes. Currently, joint point-based and voxel-based methods [12] have appeared in many detection frameworks. Discriminative features are learned through voxels and points in corresponding spatial locations. Whereas, most of the works based on combined methods [10, 12, 13] use voxel-based detection heads. Although this improves the recall of positive samples, it also comes with some drawbacks based on the voxel-based detection head, such as invalid detection positions, and presets for multiple types of anchors. This paper proposes a novel voxel-point-based combined method named SGNet, which uses a joint loss function to constrain feature learning, and designs a point-based detection head to avoid redundant calculations in the voxel-based method. And, the Voxel-To-Point Module (VTPM) is introduced into SGNet to learn the point-based semantic and geometric features of the point clouds scene.

For the two-stage object detector [14, 15, 16, 17], it is necessary to extract valuable proposals of interest from the rough prediction of the first stage. However, the misalignment between the predicted confidence and the bounding box also appears in this step. Most of the current work focuses on solving the misalignment in the second stage but ignores this in the RoI generation stage, especially when using point-based detection heads. In this paper, we propose a confidence adjustment module (CAM) with the center-boundary-aware confidence attention to solve the misalignment between the predicted confidence and proposals.

In general, to solve the learning of discriminative features in the point space based on the combined voxel and point method, and the misalignment between the predicted confidence and the bounding box when selecting RoIs, this paper proposes an end-to-end 3D detector SGNet. The main contributions can be summarized as follows.

- We propose a voxel-to-point module (VTPM) to captures semantic and geometric features for 3D object detection in the point space, which is more conducive to the detection of small-size objects.

- A confidence adjustment module (CAM) is designed for alignment between the predicted confidence and proposals in the RoI selection.

- This paper proposes a novel end-to-end two-stage 3D object detection network whose performance is competitive to the SOTA algorithm on the KITTI dataset.

The rest of the paper is organized as follows. Second II introduces some related works. The proposed SGNet model is detailed in Section III. Section IV shows experimental results and ablation studies. Finally, conclusions are presented in Section V.

# 2 Related works

## 2.1 3D object detection from point clouds

Generally, the current 3D object detection methods can be divided into two types: one-stage and two-stage detector. Single-stage detectors [18, 19, 20] regress bounding box and confidence directly, and two-stage detectors use a second stage to refine the first-stage predictions with region-proposal-aligned features. VoxelNet [7] designed a voxel feature encoding module (VFE) to extract and aggregate the point features in each voxel, which is then followed by RPN to generate detections. SECOND [8] investigates an improved sparse convolution method [21, 22, 23] based on VoxelNet, which significantly increases the speed of both training and inference. However, both methods are difficult to abandon the expensive 3D convolution. PointPillars [9] uses a novel encoder that learns features on pillars of the point cloud to predict 3D oriented boxes for objects. The two-stage detectors [14, 15, 16, 17] can usually obtain better detection results due to the refinement of the second stage. PointRCNN [3] uses Pointnet++ [6] as the backbone in both stage and devices an anchor-free strategy to generate 3D proposals. Part-$A^2$ [4] replaces the PointNet++ [6] backbone with a sparse convolutional network, and proposes RoI-aware point cloud feature pooling in the refinement. PV-RCNN [12] uses set abstraction modules to extract point features from muiti-scale voxel features in the first stage to refine the region proposals. CenterPoint [24] first detects centers of objects using a keypoint detector and regresses to other attributes. CIA-SSD [20] designs a lightweight spatial-semantic feature aggregation module to adaptively fuse high-level abstract semantic features and low-level spatial features for accurate predictions of bounding boxes and classification confidence. Voxel R-CNN [13] utilizes voxel RoI pooling to extract region features from 3D voxel features for further refinement after generating dense region proposals from bird-eye-view feature representations.

## 2.2 Voxel-point based point cloud feature learning

Object detection based on 2D image data has been widely used in daily intelligent perception. Voxel-based 3D object detection converts irregular point clouds into regular image-like data, and implements detection tasks with the help of convolutional networks. Because feature learning based on deep networks has developed very maturely, these methods are more conducive to the extraction of discriminative features. The unit of the voxel-based method is a voxel, which limits the acquisition of high-resolution geometric features. And the detection result is sensitive to the voxel size. The point-based methods [5, 6] can facilitate the learning of the geometric characteristics of the neighborhood of points. Whereas the neighborhood searching is time-consuming because it needs to traverse the candidate points. In recent years, joint point-based and voxel-based methods [10, 12] have become a new direction in 3D feature detection. The combination of features generated by different methods can be divided into the phased combination and synchronized combination. Actually, phased combinations most often appear in two-stage networks. Part-$A^2$ [4] first predicts proposals based on point features, and then uses sparse convolution to extract voxel features for two-stage bounding box refinement. STD [16] uses PointNet [5, 6] to learn point features in the first stage, and proposes PointsPool layer to obtain the representation of voxels in the proposal. There is no doubt that these methods based on point first and then voxel are not sufficient for semantic feature learning in the first stage. 3D IoU-Net [25], RangeRCNN [26], BANet [27] use voxel-based methods to roughly predict the bounding box, and the second stage incorporates point-based geometric features for fine-tuning. The above-mentioned methods of combining features phased can make up for the lack of features to a certain extent, but the disadvantages of the point-based or voxel-based methods in the first stage still exist. PV-RCNN [12] deeply integrates both 3D voxel Convolutional Neural Network (CNN) and PointNet-based set abstraction to learn more discriminative point cloud features. Voxel R-CNN [13] introduces voxel ROI pooling, which integrates spatial semantic features learned based on voxel and accelerates the PointNet [5, 6] module. SA-Det3D [28] also achieved SOTA results on KITTI based on PV-RCNN framework. HP-RPN module composed of SPConv, Auxiliary, and Keypoint branches is proposed in SIENet [29] to integrate point-based and voxel-based features. SPG [30] generates semantic points based on foreground voxels that faithfully recover the foreground regions suffering from the "missing point" problem.

## 2.3 The misalignment between confidence and bounding box

Most current 3D object detection frameworks have the issue that the predicted confidence does not align with the bounding box, that is, the bounding box with a high score does not necessarily have a higher IoU with the ground truth. STD [16] develops an IoU estimation branch to obtain the 3D IoU between the predicted box and the corresponding ground truth, and uses it to weight the classification confidence. Part-$A^2$ [4] normalizes the 3D IoU between the proposal and corresponding ground truth box as the soft label for proposal confidence refinement. Similarly, PV-RCNN [12] also uses 3D IoU guided scores in the confidence branch of the second stage. The ACA and CGE modules are proposed in 3D IoU-Net [25] to learn the IoU sensitive features from the local point cloud. And, based on the final prediction box, IoU alignment is designed to remove redundancy. CIA-SSD [20] design the IoU-aware confidence rectification module for post-processing the confidence to alleviate the misalignment between the localization accuracy and classification confidence without having an additional network stage. 3D Region of Interest (RoI) alignment is used in From Voxel to Point [31] to crop and align the features with the proposal boxes for accurately perceiving the object position.

# 3 SGNet for 3D object detection from point clouds

In this section, an end-to-end two-stage detector named SGNet which combining semantic and geometric features is presented. The architecture of SGNet is presented in figure 1. Sec. 3.1 introduces the encoder of the raw point clouds. Sec. 3.2 details the voxel-to-point module proposed in this paper. And, we described the confidence adjustment module in Sec. 3.3. In Sec. 3.4, we focus on the introduction of the refinement in second stage. Sec. 3.5 lists the items of training loss.

## 3.1 Point cloud encoder

The point clouds retain the spatial information of the scene. The point-based features learning can capture the geometric characteristics of the scene. Whereas the voxel-based method can better capture the semantic features through the convolutional networks. So, the main purpose of the encoder in this paper is to generate input data adapted to the voxel-point-based method followed. The voxelization is adopted to the scene due to the irregularity of the raw points. And, less than $T$ lidar points are reserved in each voxel. The mean value of all points in the voxel characterizes the voxel. Following PV-RCNN [12], the voxel size is set to [0.05, 0.05, 0.1] on the *X*, *Y*, *Z* axis, respectively. Due to the small voxel size, the points after sampling still have a higher resolution, and the spatial coordinate information is preserved. We define the points after downsampling as $V = \{v_i = [x_i, y_i, z_i, r_i]^T \in \mathbb{R}^4\}_{i=1\cdots N}$, where N is the number of non-empty voxels, and $v_i$ contains XYZ mean coordinates for the $i_{th}$ voxel and $r_i$ is the received mean reflectance. On the one hand, we use sparse convolution to transform the nonempty voxel representation into a high-dimensional to obtain enhanced point-wise features $F_P$. On the other hand, the SPConvNet in SECOND [8] is introduced to perform 8x downsize on the voxelized scene. And the voxels are projected to BEV to generate features $F_V$ for subsequent discriminative semantic feature learning based on 2D convolutional networks.

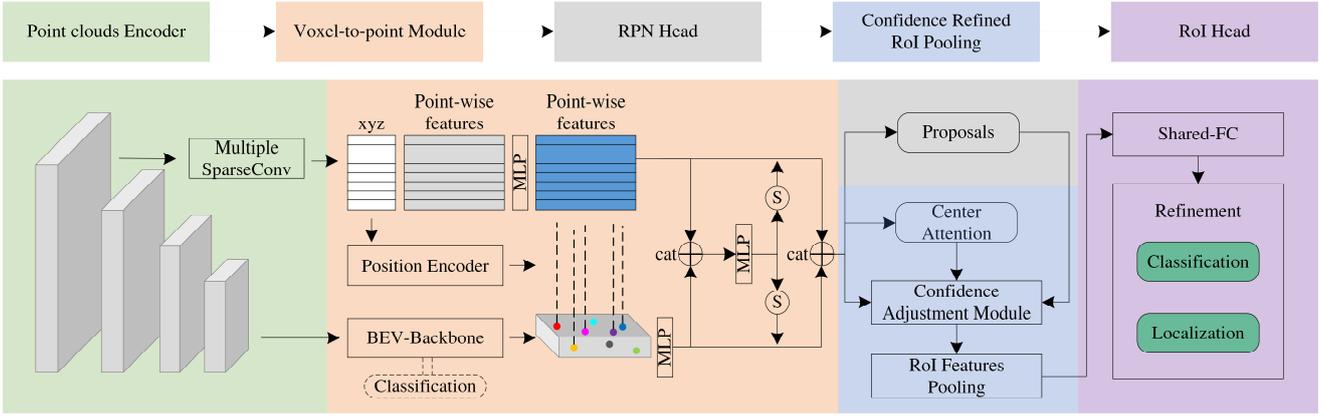

**Figure 1.** The overall architecture of our proposed SGNet.

### 3.2 Voxel-to-point module

The point-wise features $F_P$ with high resolution and voxel-wise features $F_V$ with semantic information are obtained based on the point clouds encoder. In order to augment the discrimination of features, the BEV backbone used in PointPillars and multiple Linear-BN-ReLU layers are introduced for processing of $F_P$ and $F_V$ respectively. And, the auxiliary classification task based on anchors is attached to the voxel branch for the update of parameters in the BEV backbone. The localization task is not used as an auxiliary in this paper due to the difference between the voxel-based method and the point-based method. As shown in Figure 2, for the voxel-based method, the residual between the center of the voxel and the ground truth bounding box is regressed. Whereas the point itself instead of the center is used in the point-based method. We think this would play a negative role in the localization task. On the contrary, the commonality of classification in the two methods can assist discriminative features learning. Considering the computational cost, we did not attach more auxiliary tasks.

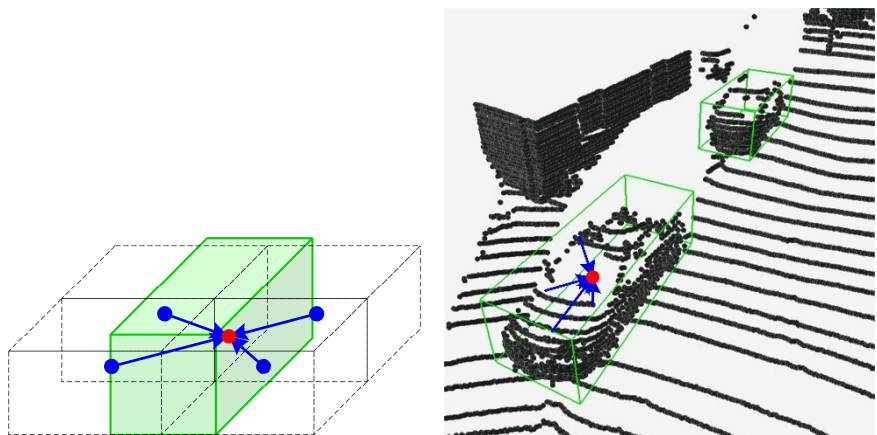

**Figure 2.** Voxel-based method (left) and point-based method (right) for localization.

The enhanced point-based geometric features and voxel-based semantic features are defined as $F_{PGeo}$ and $F_{VSem}$, respectively. In fact, the relationship between voxels and points is a one-to-many mapping. When constructing the combination of geometric and semantic features, a position encoder is introduced to make the semantic features diverse for different points. The coordinates *XYZ* of the point are encoded by multiple Liner-BN-ReLU layers to generate a position vector, which is combined with voxel-based semantic features to generate point-wise semantic features $F_{PSem}$. It is noted that the semantic features are learned through the voxel branch, while the geometric features are obtained from the point branch. Therefore, we believe that it is necessary to make soft adjustments to point-wise features due to the difference in learning methods which one is the point-based method and another is voxel-based. As shown in Figure 1, firstly, we concatenate the point-wise semantic features and geometric features to obtain the feature $F_{SG} = [F_{PSem}, F_{PGeo}]$. So $F_{SG}$ contains all the information of the two features. And, we learn the part-attention of both semantic features and the geometric features based on multiple Linear-BN-ReLU layers, which are denoted as $S_{Sem}$ and $S_{Geo}$ respectively. Finally, the combination $F_{CSG}$ of semantic and geometric features is generated by the concatenating of weighted point-wise features, as shown in Equation (1).

$$F_{CSG} = [F_{PSem} \cdot S_{Sem}, F_{PGeo} \cdot S_{Geo}] \tag{1}$$

### 3.3 Confidence adjustment module

As shown in figure 3, lidar can only scan the surface of the object. Most of the high-quality point clouds will concentrate on the boundary of large-size objects such as cars and the center of the bounding boxes for small size objects such as cyclists. Dense point clouds in local are more conducive to feature learning, that is, features are more discriminative. Actually, the classification and the localization task in the first stage are based on the features generated by the same backbone. Therefore, we can give more attention to the confidence predicted by dense points in the local neighborhood, and indirectly promote the localization task positively. This facilitates the selection of high-quality RoI in the second stage.

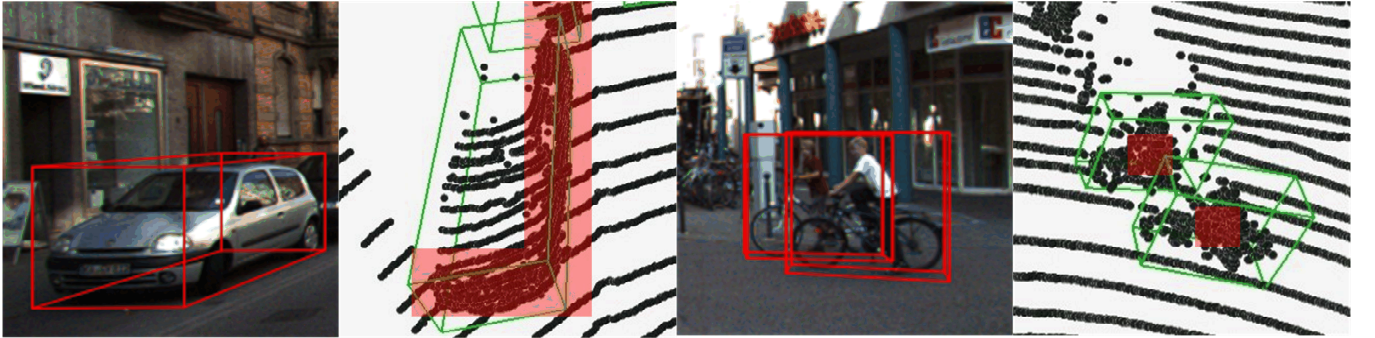

**Figure 3.** Image and point clouds representations of objects in the KITTI dataset.

We design center-boundary-aware confidence attention to drive our detector to lay emphasis on the prediction based on points close to the center and the boundary. Suppose that the ground truth bounding box is $[x, y, z, l, w, h, r]$, and an inner point $P$ is $[x_p, y_p, z_p]$. The ground truth bounding box is transformed to the local coordinates $C_{loc}$ with the center point as the origin. So, the point $P$ in this coordinate system is denoted as $[x_{loc}^p, y_{loc}^p, z_{loc}^p]$. As shown in Figure 4, take the center point as the origin to construct the attention coordinate. Decreasing attention value along with both the positive and negative directions of each coordinate axis of $C_{loc}$. When the offset is $[l/2, w/2, h/2]$, it decreases to 0, and then increases. The attention value is set to 1 When the offset relative to the center point is $[l, w, h]$.

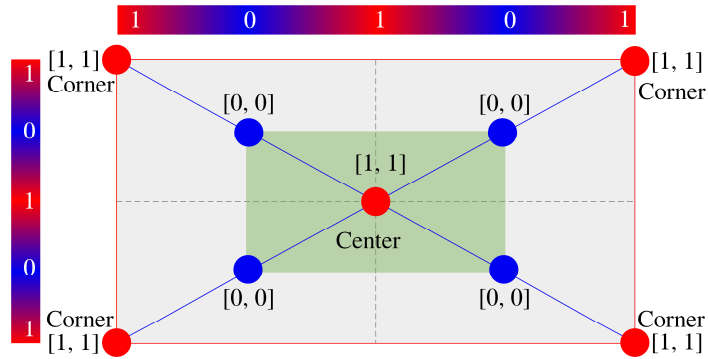

**Figure 4.** Details of confidence attention.

The entire confidence attention equation can be expressed algebraically as Equation (2).

$$CA = \left| \left( \left| \frac{[x_{loc}^p, y_{loc}^p, z_{loc}^p]}{[l, w, h]} \right| * 2 - 0.5 \right) * 2 \right| \tag{2}$$

We follow the method of normalized IoU [4]. And, the normalized confidence attention is used to the soft adjustment of confidence predicted in the first stage, as shown in Equation (3).

$$\boldsymbol{CA_{norm} = min(1, max(0, 2*CA - 0.5))} \tag{3}$$

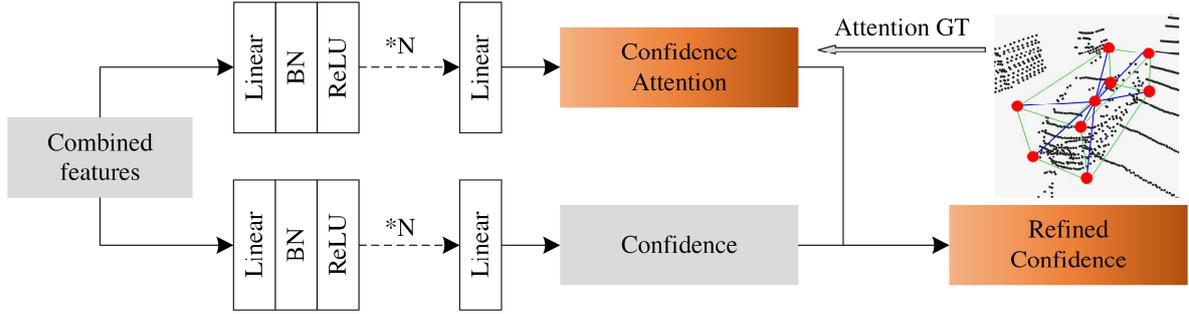

**Figure 5.** Confidence adjustment module.

As shown in figure 5, the confidence attention is derived from combined feaures through multiple Linear-BN-ReLU layers. The weighted confidence based on the mean of $CA_{norm}$ is used to select RoIs in the second stage. The Confidence Adjustment Module (CAM) introduces a center-boundary-aware confidence attention mechanism, which is conducive to uniformity of confidence, and learning of features as CAM lays more emphasis on the dense points.

### 3.4 Refinement with the combined features

The point-wise combined features which encode the semantic and geometric information are used to refine the rough results produced in the first stage. We introduce Voxel RoI Pooling Layer [13] to generate the representation of RoIs. Linear-BN-ReLU layers with shared weights transform the feature dimension into a low dimension. Then for the refinement of confidence and localization, we set two branches which consisted of multiple Linear-BN-ReLU layers to fine-tune the features, and use a Linear layer to make the final prediction. The dropout layer is adapted to prevent overfitting in the training stage. And, the 3D IoU between RoI and ground truth guided the refinement of confidence.

### 3.5 Training loss

The SGNet proposed in this paper is a two-stage network. Rough predictions of the category confidence and bounding boxes of objects in the scene are obtained in the first stage. The focal loss [32] with default parameters setting is used to overcome the class imbalance problem in classification tasks and auxiliary tasks. And, we follow the work of Part-$A^2$ [4], the residuals between ground truth and preset such as the mean size of objects, RoIs are regressed for the prediction of bounding boxes. It is noted that the corner point regularization term is introduced for localization refinement in the second stage. We use SmoothL1 loss for the regression of bounding boxes both in the first and second phases. Binary cross-entropy loss is adapted to constrain the update of the CAM network parameters, and the refinement of confidence in the second stage.

In summary, the total loss $L$ in this paper consists of six items, as shown in Equation (4). In the first stage, the auxiliary task loss $L_{aux}$, classification loss $L_{cls}$, localization loss $L_{loc}$, the CAM loss $L_{cam}$; In the second stage, confidence refinement loss $L_{cref}$, localization refinement loss $L_{lref}$.

$$L = L_{aux} + L_{cls} + L_{loc} + L_{cam} + L_{cref} + L_{lref} \tag{4}$$

## 4 Experiments

### 4.1 Datasets

The proposed detector is trained and evaluated on the widely used KITTI Object Detection Benchmark. 7481 training samples and 7518 test samples in the KITTI dataset cover cars, pedestrians, and cyclists. Each object in the scene is tagged as one of the easy, moderate, and hard levels according to the size, occlusion, and truncation of the object. The training samples are split into the *train* set (3712 samples) and *val* sets (3769 samples) for evaluation. The label in the test samples is not available. We submit the result to the KITTI website for comparing the performance of the detector proposed in this paper with other SOTA methods on the *test* set.

### 4.2 Implementation Details

For the KITTI dataset, the point clouds corresponding to the foreground image in the range of [0, 70.4]×[-40,40]×[-3, 1] along the *XYZ* axis is used in the stage of training and inference. The size of the initial voxel is [0.05, 0.05, 0.1]. At most 5 points are reserved in each voxel. And, each point is characterized by a 4-dim vector of [*x, y, z, r*]. $F_P$ and $F_V$ generated based on point clouds encoder are 64 and 256 dimensions, respectively. The MLP for $F_P$ is composed of two Subconv3D-BN-ReLU layers with [32, 64] dimensions. The BEV backbone is composed of convolutional blocks of 1x and 2x BEV scales, each of which consists of 5 Conv2d-BN-ReLU layers. The number of channels is 64, 128 respectively. Then the two scales are converted to 1x BEV size, 128-dim features through deconvolution. That is, the dimension of $F_{VSem}$ is 256. Position encoder uses the [64, 256] Linear-BN-ReLU layers to encode the position information in a 256-dimensional vector for $F_{VSem}$ enhancement, and then passes through the [128, 128] Linear-BN-ReLU layers to generate $F_{PSem}$ with 128 channels. $F_{PGeo}$ is derived from $F_V$ based on [64, 128] Linear-BN-ReLU layers. For predictions $S_{Sem}$ and $S_{Geo}$, two branches with [128, 64] Linear-BN-ReLU layers are used respectively. In the first stage, each task contains the [128, 64] LBR layers. For the auxiliary task, we set a total of 6 anchors in each voxel, with three sizes of [3.9, 1.6, 1.56], [0.8, 0.6, 1.73], [1.76, 0.6, 1.73] corresponding to cars, pedestrians, cyclists, and two angles of 0, $\pi/2$. We only set the same three sizes as above for the target category in KITTI as the regression benchmark for the first stage localization, and each point only predicts one bounding box. In the RoI pooling stage, we sample 6x6x6 positions in RoI. The search range of the Voxel RoI Pooling layer is [4, 4, 4] [8, 8, 8]. Other configurations in the second stage are the same as Voxel R-CNN [13].

The sparsity, irregularity of point clouds, and sample imbalance make the obstacles to the training of the model for detection. We use a series of augmentations to overcome this problem. The object which inner points less than 5 is abandoned in the training stage since we believe that too few points can easily add ambiguity to feature extraction. In order to make the detector robust for the irregularity of point clouds, we randomly flip the scene along the X-axis, and randomly rotate along the Z-axis a value that conforms to the uniform distribution of [-$\pi/4, \pi/4$]. Global scaling with a random factor sampled from uniformly distributed [0.95, 1.05]. More importantly, we follow the SECOND [8] and introduce the ground truth sampling augmentation to alleviate the insufficiency of ground truth samples. And, similar to PV-RCNN [12], road plane augmentation is also used in the training phase.

We trained our model with 50 epochs using Adam optimizer on NVIDIA RTX 2060 Super GPU. The batch size is set to 3. We initialize the learning rate to 0.003. The learning rate decay strategy is consistent with Voxel R-CNN. 128 RoIs are reserved for each point cloud scene. We use the class-wise score threshold to prefilter some predictions, and then the 3D IoU based NMS algorithm is adapted to remove redundant bounding boxes.

### 4.3 Evaluation on KITTI *test* dataset

In this paper, 80% of *train+val* split samples are used for training and the remaining 20% for validation. Cars, pedestrians, and cyclists use score thresholds of 0.7, 0.3, 0.3 respectively in the post-processing process. We submit the results of the test set to KITTI's official website for evaluation of our detector, and compare it with the SOTA algorithm in recent years, as shown in Tables 1 and 2. Actually, as of September 19th, our algorithm ranked **1st** in 3D and BEV detection for cyclists with easy difficulty. Ranked **2nd** in the 3D detection of moderate cyclists. In addition, SGNet proposed in this paper also achieves SOTA results on the detection of cars and pedestrians.

#### 4.3.1 Comparison of results on the 3D object detection

SGNet has achieved SOTA results on the 3D object detection, as shown in Table 1. Compared with the best results of other methods listed in Table 1, the algorithm in this paper has achieved a very large improvement in the detection of cyclists, and achieved 4.27%, 6.3%, 5.08% improvements in easy, moderate, and hard levels respectively. Compared with Voxel R-CNN, SGNet achieves 0.23% and 0.41% improvement in moderate and hard cars detection. SGNet has achieved a lead of 2.57% and 2.88% compared with the multi-modal algorithm EPNet. SGNet is superior to PV-RCNN, increasing 0.42%, 0.65% mAP for the detection with the moderate and hard cars, and achieves 8.15%, 6.69%, and 5.08% large improvement on cyclists. It is noted that SGNet has achieved the best in the moderate and hard levels compared to all the car-only detectors listed in Table 1, which shows the effectiveness and ease of training of the detector in this paper.

**Table 1** Results on the KITTI *test* 3D detection benchmark. (Mod) Moderate. (-) Not available.

| Method | Reference | Modality | Car(3D) | | | Pedestrian(3D) | | | Cyclist(3D) | | |
|---|---|---|---|---|---|---|---|---|---|---|---|
| | | | Easy | Mod. | Hard | Easy | Mod. | Hard | Easy | Mod. | Hard |
| F-PointNet[2] | CVPR2018 | RGB+Lidar | 82.19 | 69.79 | 60.59 | 50.53 | 42.15 | 38.08 | 72.27 | 56.12 | 49.01 |
| AVOD-FPN[15] | IROS2018 | RGB+Lidar | 83.07 | 71.76 | 65.73 | 50.46 | 42.47 | 39.04 | 63.76 | 50.55 | 44.93 |
| MMF[17] | CVPR2019 | RGB+Lidar | 88.40 | 77.43 | 70.22 | - | - | - | - | - | - |
| EPNet[19] | ECCV2020 | RGB+Lidar | 89.81 | 79.28 | 74.59 | - | - | - | - | - | - |
| SECOND[8] | Sensors2018 | Lidar | 83.34 | 72.55 | 65.82 | - | - | - | 71.33 | 52.08 | 45.83 |
| PointPillars[9] | CVPR2019 | Lidar | 82.58 | 74.31 | 68.99 | 51.45 | 41.92 | 38.89 | 77.10 | 58.65 | 51.92 |
| PointRCNN[3] | CVPR2019 | Lidar | 86.96 | 75.64 | 70.70 | 47.98 | 39.37 | 36.01 | 74.96 | 58.82 | 52.53 |
| STD[16] | ICCV2019 | Lidar | 87.95 | 79.71 | 75.09 | 53.29 | 42.47 | 38.35 | 78.69 | 61.59 | 55.30 |
| 3D IoU-Net[25] | Arxiv2020 | Lidar | 87.96 | 79.03 | 72.78 | - | - | - | - | - | - |
| SA-SSD[10] | CVPR2020 | Lidar | 88.75 | 79.79 | 74.16 | - | - | - | - | - | - |
| TANet[11] | AAAI2020 | Lidar | 84.39 | 75.94 | 68.82 | 53.72 | **44.34** | **40.49** | 75.70 | 59.44 | 52.53 |
| 3D-SSD[18] | CVPR2020 | Lidar | 88.36 | 79.57 | 74.55 | **54.64** | 44.27 | 40.23 | 82.48 | 64.10 | 56.90 |
| Part-$A^2$[4] | TPAMI2020 | Lidar | 87.81 | 78.49 | 73.51 | 53.10 | 43.35 | 40.06 | 79.17 | 63.52 | 56.93 |
| PV-RCNN[12] | CVPR2020 | Lidar | 90.25 | 81.43 | 76.82 | 52.17 | 43.29 | 40.29 | 78.60 | 63.71 | 57.65 |
| VoxelRCNN[13] | AAAI2021 | Lidar | **90.90** | 81.62 | 77.06 | - | - | - | - | - | - |
| CIA-SSD[20] | AAAI2021 | Lidar | 89.59 | 80.28 | 72.87 | - | - | - | - | - | - |
| SGNet(Ours) | - | Lidar | 88.83 | **81.85** | **77.47** | 49.68 | 43.00 | 40.45 | **86.75** | **70.40** | **62.73** |

Table 2 Results on the KITTI *test* BEV detection benchmark. (Mod) Moderate. (-) Not available.

| Method | Reference | Modality | Car(BEV) | | | Pedestrian(BEV) | | | Cyclist(BEV) | | |
|---|---|---|---|---|---|---|---|---|---|---|---|
| | | | Easy | Mod. | Hard | Easy | Mod. | Hard | Easy | Mod. | Hard |
| F-PointNet[2] | CVPR2018 | RGB+Lidar | 91.17 | 84.67 | 74.77 | 57.13 | 49.57 | 45.48 | 77.26 | 61.37 | 53.78 |
| AVOD-FPN[15] | IROS2018 | RGB+Lidar | 90.99 | 84.82 | 79.62 | 58.49 | 50.32 | 46.98 | 69.39 | 57.12 | 51.09 |
| MMF[17] | CVPR2019 | RGB+Lidar | 93.67 | 88.21 | 81.99 | - | - | - | - | - | - |
| EPNet[19] | ECCV2020 | RGB+Lidar | 94.22 | 88.47 | 83.69 | - | - | - | - | - | - |
| SECOND[8] | Sensors2018 | Lidar | 89.39 | 83.77 | 78.59 | - | - | - | 76.50 | 56.05 | 49.45 |
| PointPillars[9] | CVPR2019 | Lidar | 90.07 | 86.56 | 82.81 | 57.60 | 48.64 | 45.78 | 79.90 | 62.73 | 55.58 |
| PointRCNN[3] | CVPR2019 | Lidar | 92.13 | 87.39 | 82.72 | 54.77 | 46.13 | 42.84 | 82.56 | 67.24 | 60.28 |
| STD[16] | ICCV2019 | Lidar | 94.74 | 89.19 | 86.42 | 60.02 | 48.72 | 44.55 | 81.36 | 67.23 | 59.35 |
| 3D IoU-Net[25] | Arxiv2020 | Lidar | 94.76 | 88.38 | 81.93 | - | - | - | - | - | - |
| SA-SSD[10] | CVPR2020 | Lidar | **95.03** | **91.03** | 85.96 | - | - | - | - | - | - |
| TANet[11] | AAAI2020 | Lidar | 91.58 | 86.54 | 81.19 | **60.85** | **51.38** | **47.54** | 79.16 | 63.77 | 56.21 |
| 3D-SSD[18] | CVPR2020 | Lidar | 92.66 | 89.02 | 85.86 | 60.54 | 49.94 | 45.73 | 85.04 | 67.62 | 61.14 |
| Part-$A^2$[4] | TPAMI2020 | Lidar | 91.70 | 87.79 | 84.61 | 59.04 | 49.81 | 45.92 | 83.43 | 68.73 | 61.85 |
| PV-RCNN[12] | CVPR2020 | Lidar | 94.98 | 90.65 | 86.14 | 59.86 | 50.57 | 46.74 | 82.49 | 68.89 | 62.41 |
| VoxelRCNN[13] | AAAI2021 | Lidar | 94.85 | 88.83 | 86.13 | - | - | - | - | - | - |
| CIA-SSD[20] | AAAI2021 | Lidar | 93.74 | 89.84 | 82.39 | - | - | - | - | - | - |
| SGNet(Ours) | - | Lidar | 93.04 | 89.14 | **86.54** | 53.84 | 47.29 | 44.10 | **88.03** | **73.88** | **66.84** |

### 4.3.2 Comparison of results on the BEV detection

As shown in Table 2, SGNet still has significant advantages on cyclists, with 2.99%, 4.99%, and 4.43% mAP ahead in easy, moderate, and hard levels, respectively. Compared with PV-RCNN, Voxel R-CNN, and CIA-SSD, the algorithm in this paper has achieved 0.4%, 0.41%, and 4.15% mAP improvement on the detection of difficult vehicles, respectively. SGNet is superior to EPNet with a 0.67%, 2.85% increasing on moderate, hard cars respectively.

### 4.4 Evaluation on KITTI *val* dataset

Table 3 Performance comparison on KITTI *val* set with AP@11 for cars. (Mod) Moderate.

| Method | Reference | Modality | Car(3D) | | |
|---|---|---|---|---|---|
| | | | Easy | Mod. | Hard |
| F-PointNet[2] | CVPR2018 | RGB+Lidar | 83.76 | 70.92 | 63.65 |
| SECOND[8] | Sensors2018 | Lidar | 88.61 | 78.62 | 77.22 |
| PointPillars[9] | CVPR2019 | Lidar | 86.62 | 76.06 | 68.91 |
| PointRCNN[3] | CVPR2019 | Lidar | 88.88 | 78.63 | 77.38 |
| STD[16] | ICCV2019 | Lidar | 89.70 | 79.80 | 79.30 |
| 3D IoU-Net[25] | Arxiv2020 | Lidar | 89.31 | 79.26 | 78.68 |
| SA-SSD[10] | CVPR2020 | Lidar | **90.15** | 79.91 | 78.78 |
| TANet[11] | AAAI2020 | Lidar | 87.52 | 76.64 | 73.86 |
| 3D-SSD[18] | CVPR2020 | Lidar | 89.71 | 79.45 | 78.67 |
| Part-$A^2$[4] | TPAMI2020 | Lidar | 89.47 | 79.47 | 78.54 |
| PV-RCNN[12] | CVPR2020 | Lidar | 89.35 | 83.69 | 78.70 |
| Voxel R-CNN[13] | AAAI2021 | Lidar | 89.41 | 84.52 | **78.93** |
| CIA-SSD[20] | AAAI2021 | Lidar | 90.04 | 79.81 | 78.80 |
| SGNet(Ours) | - | Lidar | 89.08 | **85.10** | 78.79 |

Table 4 Performance comparison on KITTI *val* set with AP@40 for cars. (Mod) Moderate.

| Method | Reference | Modality | Car(3D) | | | Car(BEV) | | |
|---|---|---|---|---|---|---|---|---|
| | | | Easy | Mod. | Hard | Easy | Mod. | Hard |
| EPNet[19] | ECCV2020 | RGB+Lidar | 92.28 | 82.59 | 80.14 | 95.51 | 88.76 | 88.36 |
| PV-RCNN[3] | CVPR2020 | Lidar | **92.57** | 84.83 | 82.69 | **95.76** | 91.11 | 88.93 |
| Voxel R-CNN[13] | AAAI2021 | Lidar | 92.38 | **85.29** | 82.86 | 95.52 | **91.25** | 88.99 |
| SGNet(Ours) | - | Lidar | 92.16 | 85.01 | **82.99** | 95.54 | 91.22 | **89.26** |

We use the *train* set for SGNet training and take evaluation on the *val* set. As shown in Table 3, the algorithm in this paper has achieved very competitive results calculated by 11 recall positions. Compared with the other methods listed in Table 3, SGNet has achieved the best results in moderate difficulty level detection, which is ahead of Voxel R-CNN and PV-RCNN by 0.58% and 1.41% mAP respectively. Compared to CIA-SSD, our algorithm achieved a 5.29% mAP improvement on the moderate difficulty level. We also calculated mAP with 40 recall positions. And the comparison with EPNet, PV-RCNN, Voxel R-CNN is presented in Table 4. SGNet is superior to the other three methods in Table 4 in the hard difficulty level with slightly behind on easy, moderate difficulty levels.

## 4.5 Qualitative analysis

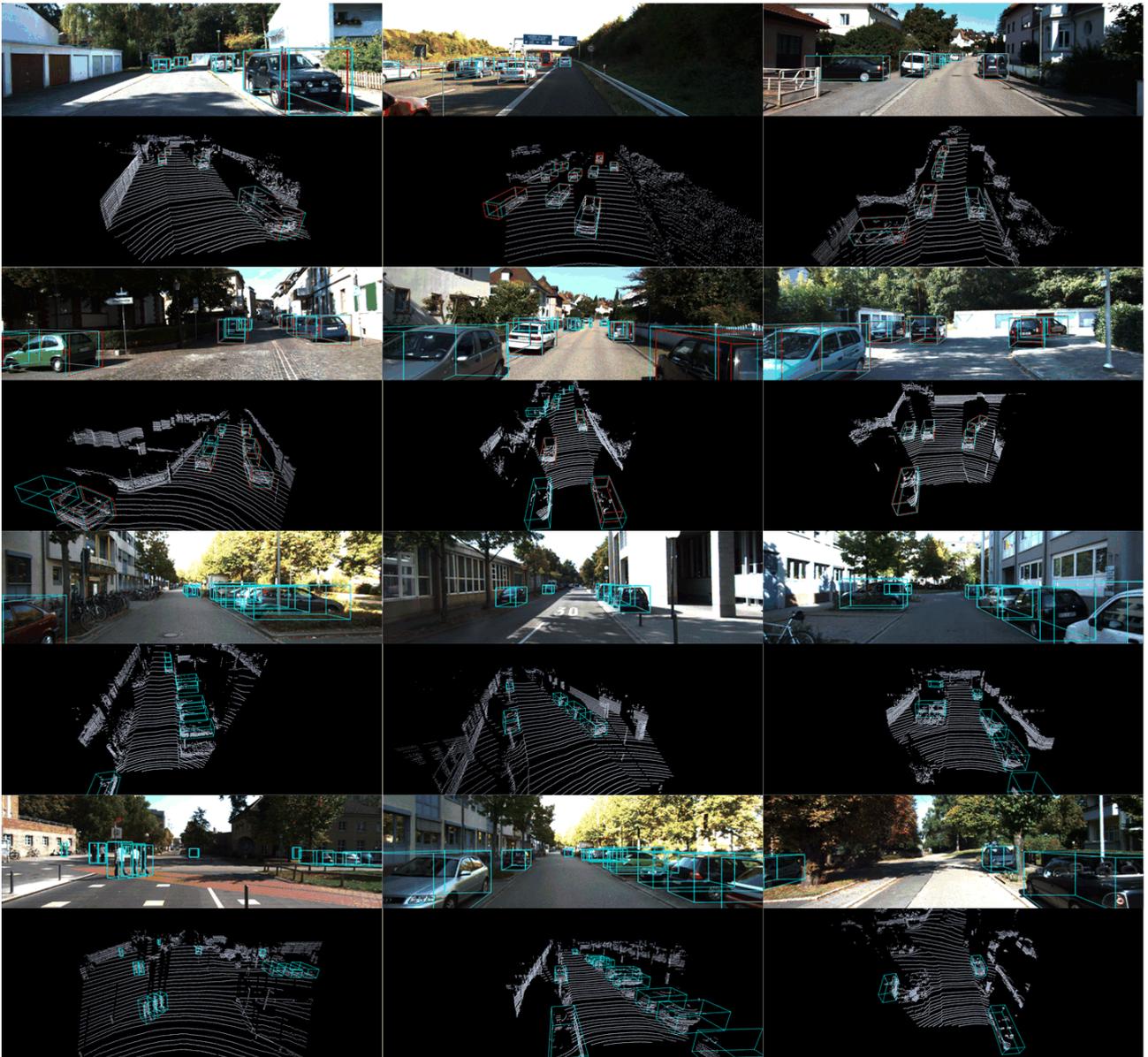

**Figure 6.** Qualitative analysis of KITTI results. The 3D detection results of our model on *val* split (top 4 rows) and *test* split (last 4 rows).The ground truth 3D bounding boxes are in red while cyan implies the result predicted by SGNet.

The representative detection results of SGNet proposed in this paper on the *val* split and *test* split are shown in Figure 6. Our model achieves good performance, and many unlabeled objects in KITTI dataset are still perceived.

## 4.6 Ablation Studies

### 4.6.1 CAM

Table 5 **Performance of CAM with AP@40 for cars. (Mod) Moderate.**

| CAM | Car(3D) | | |
|---|---|---|---|
| | Easy | Mod. | Hard |
| ✗ | 91.98 | 83.21, | 81.00 |
| ✓ | **92.16** | **85.01** | **82.99** |

In order to verify that the CAM module proposed in this paper can make contribution to the final mAP of 3D detection. We conducted a comparative experiment between the algorithm in this paper and the algorithm without CAM. We calculate mAP by 40 recall positions. The results are presented in Table 5. As shown in Table 5, that The CAM promote the performance of our algorithm on easy, moderate, hard cars detection with 0.18%, 1.8%, 1.99% mAP improvement respectively. The CAM module proposed in this paper adjusts the confidence and aligns it with the localization task. Actually, the CAM module can effectively perceive the center and boundary of the object, and perfectly predicts the confidence attention. Some instances of point cloud objects with confidence attention colored can be seen in Figure 7.

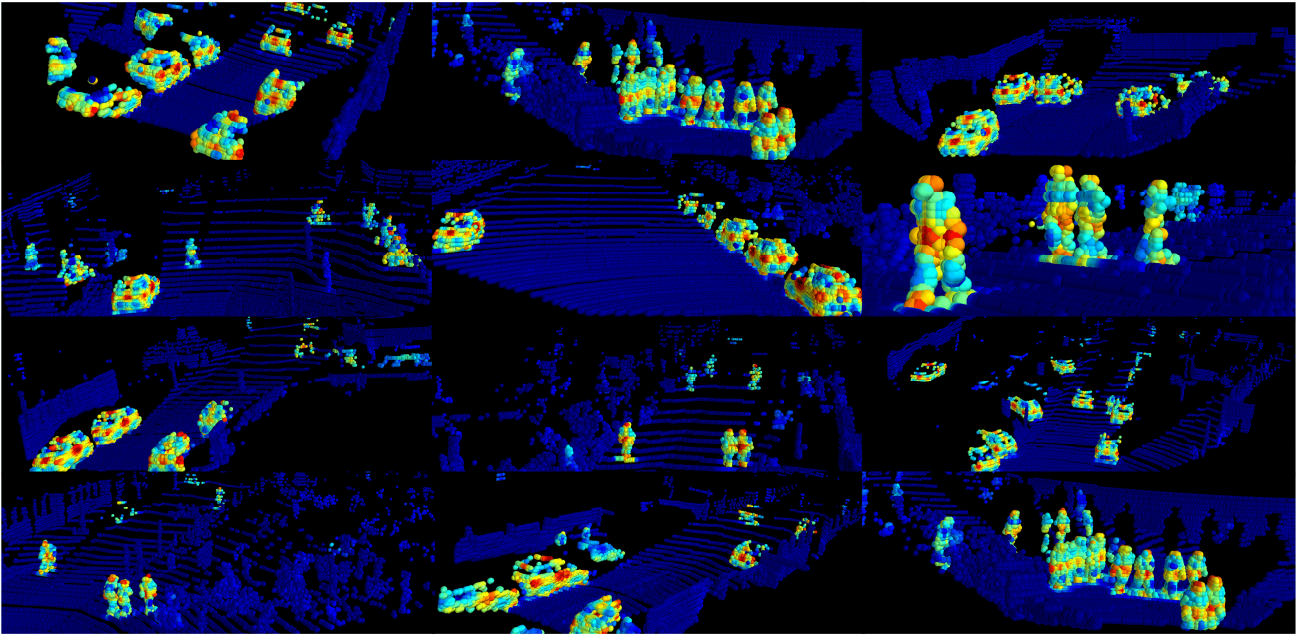

**Figure 7.** Instances of point cloud objects with confidence attention colored.

### 4.6.2 Position encoder

**Table 6** Performance of position encoder with AP@40 for cars. (Mod) Moderate.

| Position encoder | Car(3D) | | |
| --- | --- | --- | --- |
| | Easy | Mod. | Hard |
| ✕ | **92.63** | 83.25 | 82.86 |
| ✓ | 92.16 | **85.01** | **82.99** |

We also explore the impact of the position encoder module in the VTPM on 3D detection. As shown in Table 6, in order to verify the efficiency of the position encoder module, the comparison experiments which include the position encoder and another one do not are set, respectively. The results indicate that the position encoder enhances the mAP of 3D cars detection, and the promotion of 1.76%, 0.13% is obtained in moderate, hard difficulty levels, correspondingly. This also verifies our ideas, that is, increase the diversity of features by position encoding mechanisms, thereby driving the voxel-point-based 3D object detection.

## 5 Conclusion

We propose a novel end-to-end two-stage 3D object detector named SGNet that combines semantic and geometric features. The voxel-to-point module is used to construct semantic-geometric features of point clouds. And, we design the CAM based on the center-boundary-aware confidence attention for alignment between predicted confidence and proposals. The SGNet proposed in this paper has achieved SOTA performance in the KITTI dataset. However, our algorithm is not particularly strong on pedestrian detection, which may be that there are many objects similar to those in point clouds data format, such as poles. In subsequent work, we will try to resolve this issue.

## Competing interests

## Grant information

**References**
[1] A. Geiger, P. Lenz, C. Stiller, R. Urtasun. Vision meets robotics: The KITTI dataset. The International Journal of Robotics Research. 32 (2013) 1231-1237. https://doi.org/10.1177/0278364913491297.
[2] C. R. Qi, W. Liu, C. Wu, H. Su, L. J. Guibas. Frustum PointNets for 3D Object Detection from RGB-D Data. 2018 IEEE/CVF Conference on Computer Vision and Pattern Recognition (CVPR). (2018) 918-927. https://doi.org/10.1109/CVPR.2018.00102.
[3] S. Shi, X. Wang, H. Li. PointRCNN: 3D Object Proposal Generation and Detection From Point Cloud. 2019 IEEE/CVF Conference on Computer Vision and Pattern Recognition (CVPR). (2019) 770-779. https://doi.org/10.1109/CVPR.2019.00086.
[4] S. Shi, Z. Wang, J. Shi, X. Wang, H. Li. From Points to Parts: 3D Object Detection from Point Cloud with Part-aware and Part-aggregation Network. IEEE Transactions on Pattern Analysis and Machine Intelligence. https://doi.org/10.1109/TPAMI.2020.2977026.
[5] R. Q. Charles, H. Su, M. Kaichun, L. J. Guibas. PointNet: Deep Learning on Point Sets for 3D Classification and Segmentation. 2017 IEEE Conference on Computer Vision and Pattern Recognition (CVPR). (2017) 77-85. https://doi.org/10.1109/CVPR.2017.16.
[6] R. Q. Charles, H. Su, L. Yi, L. J. Guibas. PointNet++: Deep Hierarchical Feature Learning on Point Sets in a Metric Space. arXiv preprint. (2017). https://arxiv.org/abs/1706.02413.
[7] Y. Zhou, O. Tuzel. VoxelNet: End-to-End Learning for Point Cloud Based 3D Object Detection. 2018 IEEE/CVF Conference on Computer Vision and Pattern Recognition. (2018) 4490-4499. https://doi.org/10.1109/CVPR.2018.00472.
[8] Y. Yan, Y. Mao, B. Li. SECOND: Sparsely Embedded Convolutional Detection. Sensors. 18 (2018) 3337. https://doi.org/10.3390/s18103337.
[9] A. H. Lang, S. Vora, H. Caesar, L. Zhou, J. Yang, O. Beijbom. PointPillars: Fast Encoders for Object Detection From Point Clouds. 2019 IEEE/CVF Conference on Computer Vision and Pattern Recognition (CVPR). (2019) 12689-12697. https://doi.org/10.1109/CVPR.2019.01298.
[10] C. He, H. Zeng, J. Huang, X.-S. Hua, L. Zhang. Structure Aware Single-Stage 3D Object Detection From Point Cloud. 2020 IEEE/CVF Conference on Computer Vision and Pattern Recognition (CVPR). (2020) 11870-11879. https://doi.org/10.1109/CVPR42600.2020.01189.
[11] Z. Liu, X. Zhao, T. Huang, R. Hu, Y. Zhou, X. Bai. TANet: Robust 3D Object Detection from Point Clouds with Triple Attention. Proceedings of the AAAI Conference on Artificial Intelligence. 34 (2020) 11677-11684. https://doi.org/10.1609/aaai.v34i07.6837.
[12] S. Shi, C. Guo, L. Jiang, Z. Wang, J. Shi, X. Wang, H. Li. PV-RCNN: Point-Voxel Feature Set Abstraction for 3D Object Detection. 2020 IEEE/CVF Conference on Computer Vision and Pattern Recognition (CVPR). (2020) 10526-10535.
[13] J. Deng, S. Shi, P. Li, W. Zhou, Y. Zhang, H. Li. Voxel R-CNN: Towards High Performance Voxel-based 3D Object Detection. AAAI2021. https://arxiv.org/abs/2012.15712v2.
[14] D. Zhou, J. Fang, X. Song, L. L, J. Y, Y. Dai, H. Li, R. Yang. Joint 3D Instance Segmentation and Object Detection for Autonomous Driving. 2020 IEEE/CVF Conference on Computer Vision and Pattern Recognition (CVPR). (2020) 1836-1846. https://doi.org/10.1109/CVPR42600.2020.00191.
[15] J. Ku, M. Mozifian, J. Lee, A. Harakeh, S. L. Waslander. Joint 3D Proposal Generation and Object Detection from View Aggregation. 2018 IEEE/RSJ International Conference on Intelligent Robots and Systems (IROS). (2018) 1-8. https://doi.org/10.1109/IROS.2018.8594049.
[16] Z. Yang, Y. Sun, S. Liu, X. Shen, J. Jia. STD: Sparse-to-Dense 3D Object Detector for Point Cloud. 2019 IEEE/CVF International Conference on Computer Vision (ICCV). (2019) 1951-1960. https://doi.org/10.1109/ICCV.2019.00204.
[17] M. Liang, B. Yang, Y. Chen, R. Hu, R. Urtasun. Multi-Task Multi-Sensor Fusion for 3D Object Detection. 2019 IEEE/CVF Conference on Computer Vision and Pattern Recognition (CVPR). (2019) 7337-7345. https://doi.org/10.1109/CVPR.2019.00752.
[18] Z. Yang, Y. Sun, S. Liu, J. Jia. 3DSSD: Point-Based 3D Single Stage Object Detector. 2020 IEEE/CVF Conference on Computer Vision and Pattern Recognition (CVPR). (2020) 11037-11045. https://doi.org/10.1109/CVPR42600.2020.01105.
[19] T. Huang, Z. Liu, X. Chen, X. Bai. EPNet: Enhancing Point Features with Image Semantics for 3D Object Detection. ECCV2020. https://arxiv.org/abs/2007.08856.
[20] W. Zheng, W. Tang, S. Chen, L. Jiang, C. Fu. CIA-SSD: Confident IoU-Aware Single-Stage Object Detector From Point Cloud. AAAI2021. https://arxiv.org/abs/2012.03015.
[21] B. Graham. Sparse 3d convolutional neural networks. BMVC. (2015). https://doi.org/10.5244/C.29.150.
[22]. B. Graham and L. van der Maaten. Submanifold sparse convolutional networks. CoRR. (2017). https://arxiv.org/abs/1706.01307.
[23] B. Graham, M. Engelcke and L. v. d. Maaten. 3D Semantic Segmentation with Submanifold Sparse Convolutional Networks. 2018 IEEE/CVF Conference on Computer Vision and Pattern Recognition. (2018) 9224-9232. https://doi.org

/10.1109/ CVPR.2018.00961.

[24] T. Yin, X. Zhou, P. Krähenbühl. Center-based 3D Object Detection and Tracking. CVPR2021.

[25] J. Li, S. Luo, Z. Zhu, H. Dai, A. S. Krylov, Y. Ding, L. Shao. 3D IoU-Net: IoU Guided 3D Object Detector for Point Clouds. arXiv preprint. (2020). https://arxiv.org/abs/2004.04962.

[26] Liang, Zhidong and Zhang, Ming and Zhang, Zehan and Zhao, Xian and Pu, Shiliang. RangeRCNN: Towards Fast and Accurate 3D Object Detection with Range Image Representation. 2020

[27] R. Qian, X. Lai, X. Li. Boundary-Aware 3D Object Detection from Point Clouds. arXiv. 2021. https://arxiv.org/abs/ 2104. 10330

[28] P. Bhattacharyya, C. Huang and K. Czarnecki: SA-Det3D: Self-Attention Based Context-Aware 3D Object Detection. 2021. https://arxiv.org/abs/2101.02672.

[29] Z. Li, Y. Yao, Z. Quan, W. Yang and J. Xie: SIENet: Spatial Information Enhancement Network for 3D Object Detection from Point Cloud. 2021. https://arxiv.org/abs/2103.15396.

[30] Q. Xu, Y. Zhou, W. Wang, C. Qi and D. Anguelov: SPG: Unsupervised Domain Adaptation for 3D Object Detection via Semantic Point Generation. Proceedings of the IEEE conference on computer vision and pattern recognition (ICCV) 2021.

[31] J. Li, H. Dai, L. Shao and Y. Ding: From Voxel to Point: IoU-guided 3D Object Detection for Point Cloud with Voxel-to-Point Decoder. MM '21: The 29th ACM International Conference on Multimedia (ACM MM) 2021.

[32] T. Lin, P. Goyal, R. Girshick, K. He, P. Dollár. Focal Loss for Dense Object Detection. IEEE Transactions on Pattern Analysis and Machine Intelligence. (2020) 318-327. https://doi.org/10.1109/TPAMI.2018.2858826.